\documentclass{article}

\usepackage[preprint]{neurips_2025}

\usepackage[utf8]{inputenc} % allow utf-8 input
\usepackage[T1]{fontenc}    % use 8-bit T1 fonts
\usepackage{amsmath}
\usepackage{amsthm}

\usepackage{hyperref}       % hyperlinks
\usepackage{url}            % simple URL typesetting
\usepackage{booktabs}       % professional-quality tables
\usepackage{amsfonts}       % blackboard math symbols
\usepackage{nicefrac}       % compact symbols for 1/2, etc.
\usepackage{microtype}      % microtypography
\usepackage{xcolor}         % colors
\usepackage{graphicx}
\usepackage{diagbox}
\usepackage{url}
\usepackage{booktabs}
\usepackage{amssymb}
\usepackage{bbding}
\usepackage{pifont}
\usepackage{wasysym}
\usepackage{utfsym}
\usepackage{fontawesome}
\usepackage{appendix}
\usepackage{multirow}
\usepackage{wrapfig}

\title{FAD: Frequency Adaptation and Diversion \\for Cross-domain Few-shot Learning}

\author{%
  Ruixiao Shi\textsuperscript{\rm 1,2}\thanks{Equal contribution}\quad Fu Feng\textsuperscript{\rm 1,2}\footnotemark[1]\quad Yucheng Xie\textsuperscript{\rm 1,2}\quad Jing Wang\textsuperscript{\rm 1,2}\thanks{Corresponding authors}\quad Xin Geng\textsuperscript{\rm 1,2}\footnotemark[2] \\
  School of Computer Science and Engineering, Southeast University, Nanjing, China\\
  Key Laboratory of New Generation Artificial Intelligence Technology and Its Interdisciplinary \\Applications (Southeast University), Ministry of Education, China \\
  \texttt{\{eric\_xiao, fufeng, xieyc, wangjing91, xgeng\}@seu.edu.cn} \\
}

\begin{document}

\maketitle

\vspace{-0.15in}
\begin{abstract}
Cross-domain few-shot learning (CD-FSL) requires models to generalize from limited labeled samples under significant distribution shifts.
While recent methods enhance adaptability through lightweight task-specific modules, they operate solely in the spatial domain and overlook frequency-specific variations that are often critical for robust transfer.
We observe that spatially similar images across domains can differ substantially in their spectral representations, with low and high frequencies capturing complementary semantic information at coarse and fine levels.
This indicates that uniform spatial adaptation may overlook these spectral distinctions, thus constraining generalization.
To address this, we introduce Frequency Adaptation and Diversion (FAD), a frequency-aware framework that explicitly models and modulates spectral components.
At its core is the Frequency Diversion Adapter, which transforms intermediate features into the frequency domain using the discrete Fourier transform (DFT), partitions them into low, mid, and high-frequency bands via radial masks, and reconstructs each band using inverse DFT (IDFT).
Each frequency band is then adapted using a dedicated convolutional branch with a kernel size tailored to its spectral scale, enabling targeted and disentangled adaptation across frequencies.
Extensive experiments on the Meta-Dataset benchmark demonstrate that FAD consistently outperforms state-of-the-art methods on both seen and unseen domains, validating the utility of frequency-domain representations and band-wise adaptation for improving generalization in CD-FSL.
\end{abstract}

\vspace{-0.15in}
\section{Introduction}
\label{sec:intro}
Pre-trained models have demonstrated strong performance on diverse vision benchmarks~\cite{liu2021swin, wu2021cvt, feng2024wave}, but their generalization ability degrades under limited supervision and substantial domain shifts~\cite{feng2024transferring, kundu2022balancing, feng2023genes}. 
This limitation motivates Cross-Domain Few-Shot Learning (CD-FSL)~\cite{chen2019closer, guo2020broader}, which targets robust generalization across distributional gaps using only a few labeled samples.

CD-FSL typically involves a two-phase paradigm: a pre-training stage on source domains followed by meta-testing on unseen target domains. 
Recent work~\cite{li2022cross} demonstrates that incorporating task-specific parameters during meta-testing substantially improves generalization. 
Early approaches~\cite{Requeima2019, bateni2022enhancing} generate these parameters via auxiliary networks conditioned on task embeddings. 
In contrast, recent approaches inspired by parameter-efficient fine-tuning (PEFT)~\cite{houlsby2019parameter, lester2021power, xie2024fine} directly insert lightweight, task-adaptive modules into the pre-trained backbone~\cite{liuuniversal, li2022cross, guo2023task, wu2024task}.
These modules commonly adopt linear modulation, such as linear layers~\cite{liuuniversal} or residual 1$\times$1 convolutions~\cite{li2022cross}, which balances adaptation capacity with efficiency by reducing both overfitting and computational overhead.

\begin{figure}[t]
\centering
\includegraphics[width=1\textwidth]{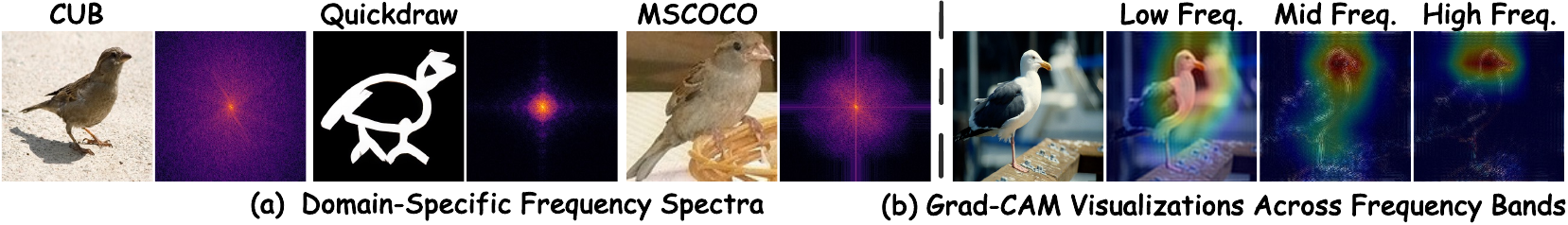}
\vspace{-0.25in}
\caption{(a)~Images with similar spatial semantics from different domains can differ significantly in their frequency spectra, reflecting domain-specific variations in texture, style, and structure.
(b)~Grad-CAM~\cite{selvaraju2017grad} visualizations with frequency masking reveal a shift in attention from coarse to fine regions as frequency increases, highlighting the semantic specificity of spectral bands.} 
\vspace{-0.15in}
\label{fig:moti}
\end{figure}

While recent approaches have achieved notable progress in CD-FSL, they predominantly focus on spatial-domain adaptation~\cite{li2022cross, guo2023task}. 
However, spatial features can overlook important differences between domains that are more clearly revealed in the frequency domain.
As illustrated in Fig.~\ref{fig:moti}a, images from different domains may appear semantically similar in the spatial domain, but their spectral distributions reveal substantially greater divergence due to differences in texture, imaging style, and background complexity—factors more effectively characterized in the frequency domain~\cite{fu2022wave, zhong2022detecting, zhou2024meta}.
This observation highlights the potential of spectral modeling for uncovering latent domain-specific variations.

Motivated by this, we explore frequency-domain representations as a means to enhance the generalization of pre-trained models under distribution shift. 
Frequency components naturally decompose visual signals into multi-scale structures, where low frequencies capture global shape and layout while high frequencies encode fine-grained textures and edges~\cite{xu2020learning, fu2022wave, liu2024continual}.
These complementary signals provide a more structured and semantically aligned basis for modeling domain shifts.
To assess their impact, we apply band-pass filters and visualize Grad-CAM~\cite{selvaraju2017grad} activations under different frequency bands. 
As shown in Fig.~\ref{fig:moti}b, model attention progressively shifts from coarse semantic regions to localized details with increasing frequency, suggesting that each spectral band contributes distinct semantics critical for cross-domain adaptation.

Such evidence reveals a key limitation in existing frequency-aware methods, which often apply identical modulation across all spectral components. 
This uniform treatment fails to account for the distinct semantic contributions of different frequency bands~\cite{huang2021fsdr, tong2024lightweight}, potentially limiting the model’s ability to adapt to domain-specific variations.
To address this, we draw on the principle of Knowledge Diversion~\cite{xie2024kind}, which enhances generalization by assigning separate parameter subsets to distinct forms of knowledge. 
Extending this principle to the spectral domain, we introduce Frequency Diversion, a mechanism that treats different frequency bands as semantically distinct components and applies dedicated parameter branches to modulate each one independently. This design allows the model to respond more precisely to the unique characteristics encoded at each frequency scale.

Building on this mechanism, we introduce Frequency Adaptation and Diversion (FAD), a frequency-aware framework tailored for CD-FSL.
FAD operates during meta-testing and incorporates multiple spectral branches to capture frequency-specific variations.
At the core of FAD is the Frequency Diversion Adapter, a lightweight residual module that transforms features into the frequency domain via discrete Fourier transform and partitions them into predefined bands using spectral masks. 
Each band is modulated by a dedicated convolutional branch, with kernel sizes aligned to its spectral granularity. 
This plug-and-play design supports fine-grained, frequency-aware adaptation and can be seamlessly integrated into existing architectures to improve generalization under domain shift.

Our key contributions are as follows:
(1) We propose Frequency Diversion, a spectral extension of knowledge diversion that enables semantically disentangled, frequency-aware adaptation for cross-domain few-shot learning.
(2) We develop the Frequency Diversion Adapter, a lightweight, modular component that performs efficient, band-specific modulation during meta-testing. It is plug-and-play and architecture-agnostic, making it compatible with existing adapter-based frameworks.
(3) We establish new state-of-the-art performance on the Meta-Dataset benchmark, demonstrating substantial improvements in few-shot generalization, particularly on challenging unseen domains.

\section{Related Work}
\subsection{Cross-domain Few-shot Learning}
Methods for CD-FSL can be broadly divided into training-time and adaptation-time approaches~\cite{guo2020broader, luo2023closer, xue2024towards, li2025svasp}. Training-time methods, typically grounded in meta-learning, aim to learn generalizable representations~\cite{liuuniversal, tian2024mokd, li2025svasp}. In contrast, adaptation-time methods operate during meta-testing and perform task-specific customization using the support set~\cite{Requeima2019, li2022cross, guo2023task, yang2024leveraging}. 
A prominent line of adaptation-time research has demonstrated strong performance by integrating lightweight task-specific modules—such as FiLM layers~\cite{Requeima2019}, linear layers~\cite{liuuniversal}, 1$\times$1 convolutions~\cite{li2022cross}, and group convolutions with normalization~\cite{yang2024leveraging}—into pre-trained backbones and updating them during meta-testing. While effective, these methods operate solely in the spatial domain and rely on linear modulation, overlooking frequency-specific variations that capture important domain- and class-dependent information. Thus, we propose a Frequency Diversion Adapter that explicitly decomposes and modulates features across distinct frequency bands to improve cross-domain adaptation.

\subsection{Frequency Domain Learning}
Frequency-domain techniques have played a foundational role in vision systems, with early work grounded in spectral analysis~\cite{marr1980theory, canny1986computational}. Recent advances have integrated these principles into deep learning, demonstrating success in semantic segmentation~\cite{bo2025famnet, tong2024lightweight}, domain generalization~\cite{zhao2022test, huang2021fsdr}, and geometric transformation tasks~\cite{xiao2024towards}.
Recent studies have highlighted the close connection between frequency components and domain-specific variations, underscoring the utility of frequency-aware modeling in cross-domain scenarios.
Methods such as DFF~\cite{lin2023deep} and CACPA~\cite{tong2024lightweight} leverage frequency-domain filters to improve performance in domain-shifted tasks, while approaches like Wave-SAN~\cite{fu2022wave} utilize frequency-based data augmentation to enhance robustness.
Other efforts~\cite{cheng2023frequency, zhou2024meta} decompose input images into spectral bands and introduce multi-branch architectures to capture frequency-specific features, aiming to improve generalization across domains.
While these methods have explored the utility of frequency-domain representations for cross-domain tasks, they typically overlook the need for targeted modulation across frequency bands. By treating all spectral components uniformly, they fail to account for the varying semantic granularity encoded at different frequencies, limiting the potential benefits of frequency-aware adaptation.

\section{Methods}
\subsection{Preliminaries}
\paragraph{Problem Formulation.}
Cross-domain few-shot learning (CD-FSL) aims to recognize novel classes in a target domain $\mathcal{X}_t$ using only a few labeled examples, where the model is pre-trained on a distinct source domain $\mathcal{X}_s$ with $\mathcal{X}_s \neq \mathcal{X}_t$. 
Each evaluation task is represented as an episode $\mathcal{T} = (\mathcal{S}, \mathcal{Q})$, where the support set $\mathcal{S} = \{(x_i, y_i)\}_{i=1}^{N\times K}$ contains $K$ labeled examples for each of $N$ classes, and the query set $\mathcal{Q} = \{x_i\}_{i=1}^{|\mathcal{Q}|}$ includes unlabeled samples from the same classes. The primary challenge lies in achieving robust generalization across domains under limited supervision.

\paragraph{Frequency Domain Transformation.}
We utilize the 2D Discrete Fourier Transform (DFT) to convert feature maps from the spatial domain to the frequency domain. 
This transformation enables explicit access to frequency components, facilitating the analysis and modulation of structural patterns that are often less distinguishable in spatial representations.

Given a spatial feature map $h \in \mathbb{R}^{W \times H \times C}$,  the DFT is applied independently to each channel, producing a complex-valued frequency representation:
\begin{equation}
    \mathcal{F}(u, v) = \sum_{m=0}^{H-1} \sum_{n=0}^{W-1} h(m, n) \cdot e^{-2\pi i \left( \frac{um}{H} + \frac{vn}{W} \right)},
\label{eq:DFT}
\end{equation}
where $h(m, n)$ denotes the spatial feature at location $(m, n)$, and $\mathcal{F}(u, v)$ is the corresponding frequency-domain coefficient at position $(u, v)$.

To convert back to the spatial domain, the inverse DFT (IDFT) is applied:
\begin{equation}
    h(x, y) = \frac{1}{WH} \sum_{u=0}^{W-1} \sum_{v=0}^{H-1} \mathcal{F}(u, v) \cdot e^{2\pi i \left( \frac{ux}{W} + \frac{vy}{H} \right)},
\label{eq:IDFT}
\end{equation}
where $h(x,y)$ denotes the reconstructed spatial feature. Both DFT and IDFT are performed per channel and independently across the feature dimensions.

\subsection{Overview of Frequency-Diversion Adapter}
Cross-domain few-shot learning typically follows a two-stage paradigm: a pre-training stage, where a model $f_{\phi}$ is trained on base classes from a source domain, and a meta-testing stage, where the model is adapted to novel classes in a target domain using only a small support set $\mathcal{S}$.
To mitigate overfitting under limited supervision, recent approaches freeze the backbone $f_{\phi}$ and introduce lightweight task-specific modules $f_{\alpha}$ for adaptation.

Following prior work~\cite{li2022cross}, we employ residual adapters inserted at various layers of the backbone. The output of the $l$-th layer is defined as:
\begin{equation}
    f_l(h) = f_{\phi_l}(h) + f_{\alpha_l}(h),
\end{equation}
where $h$ is the input feature to layer $l$ and $f_{\alpha_l}$ denotes the learnable adapter that enables parameter-efficient adaptation during meta-testing.

To more effectively exploit spectral information critical for cross-domain generalization, we introduce the Frequency-Diversion Adapter, a frequency-aware module designed for fine-grained adaptation.
The Frequency-Diversion Adapter comprises two key components: Frequency Diversion (Section~\ref{sec:FD}) and Band-wise Adaptation (Section~\ref{sec:BW}). 
\begin{itemize}
    \item \textbf{Frequency Diversion} projects feature maps into the frequency domain using the discrete Fourier transform (DFT) and partitions them into low, mid, and high-frequency bands via predefined radial masks.
    \item \textbf{Band-wise Adaptation} then reconstructs each band in the spatial domain and applies modulation through dedicated convolutional branches, enabling frequency-aware adaptation aligned with domain-specific semantic granularity.
\end{itemize}

\begin{figure}[!t]
\centering
\includegraphics[width=\textwidth]{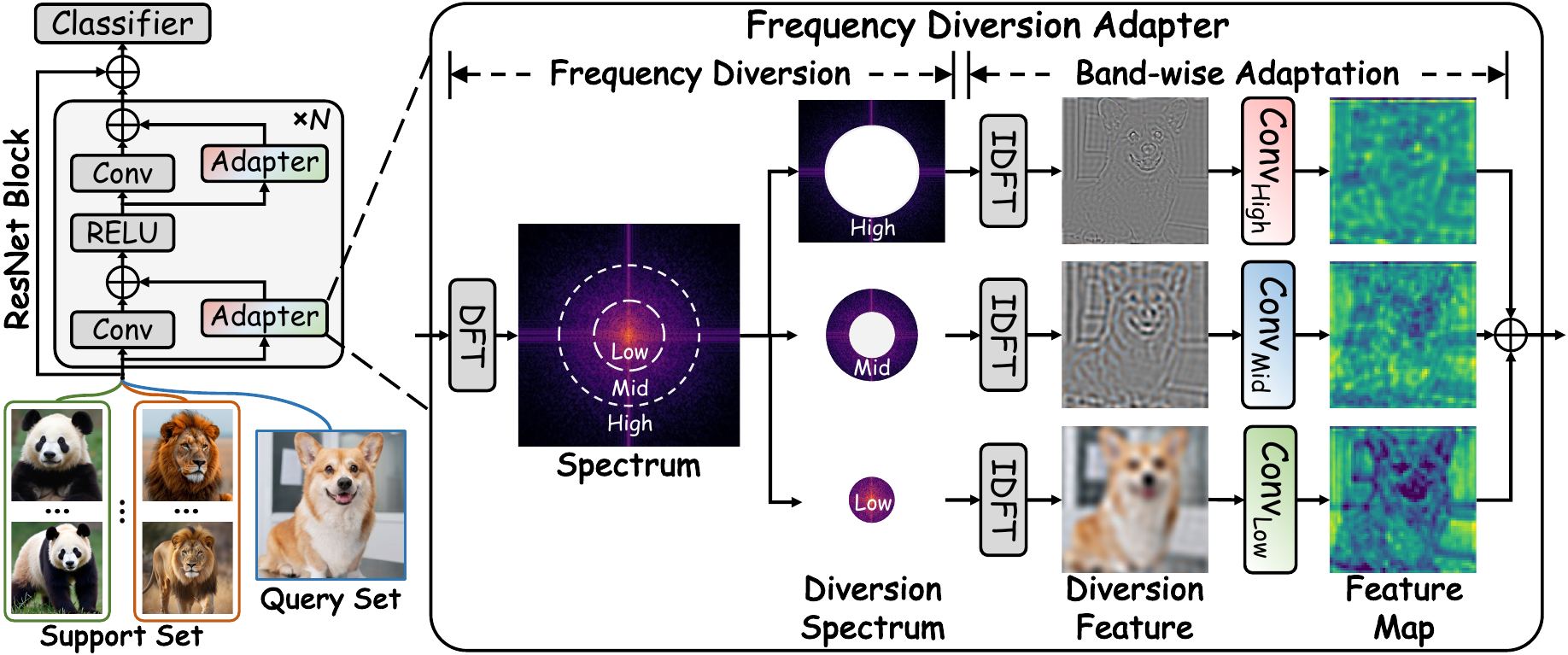}
\vspace{-0.15in}
\caption{Overview of the Frequency Diversion Adapter. \textbf{(a)~Frequency Diversion} transforms features into the frequency domain and partitions them into spectral bands using radial masks. \textbf{(b)~Band-wise Adaptation} reconstructs each band via inverse DFT and modulates it using dedicated convolutional branches. This design enables fine-grained, frequency-aware adaptation for improved generalization across domains.}
\vspace{-0.15in}
\label{fig:method}
\end{figure}

\subsection{Frequency Diversion} 
\label{sec:FD}
Frequency Diversion builds on the principle of Knowledge Diversion from KIND~\cite{xie2024kind}, which allocates disjoint parameter subsets to specialize in distinct types of knowledge. 
While originally applied to semantic decomposition in generative modeling, we extend this idea to the frequency domain, where visual features are inherently organized by granularity: low frequencies encode global structure, and high frequencies capture fine textures and edges.

This spectral hierarchy offers a principled basis for domain-aware adaptation. By partitioning feature maps into low, mid, and high-frequency bands and modulating each through a dedicated parameter branch, Frequency Diversion enables semantically disentangled adaptation across spectral scales, enhancing the model’s capacity to capture domain-specific variation.

Given an input feature map $h \in \mathbb{R}^{W \times H \times C}$, we apply the 2D Discrete Fourier Transform (DFT) independently across channels to obtain its frequency-domain representation $\mathcal{F}\in \mathbb{C}^{W \times H \times C}$, as defined in Eq.~\eqref{eq:DFT}.
To divide the frequency spectrum, we define radial masks based on the normalized Euclidean distance from the spectral center $(u_0, v_0) = (\frac{H}{2}, \frac{W}{2})$:
\begin{equation}
    M^{(r_1, r_2)}(u, v) = 
    \begin{cases}
    1, & \text{if } r_1 < \sqrt{\left(\frac{u - u_0}{u_0}\right)^2 + \left(\frac{v - v_0}{v_0}\right)^2} \leq r_2 \\
    0, & \text{otherwise}
    \end{cases}
\end{equation}
where $r_1, r_2\in [0,1]$ define the inner and outer boundaries of the spectral band.

The masked frequency components are computed as:
\begin{equation}
\mathcal{F}^{(r_1, r_2)}(u, v) = M^{(r_1, r_2)}(u, v) \cdot \mathcal{F}(u, v)
\end{equation}
In practice, we define three disjoint bands to isolate low-, mid-, and high-frequency components, denoted as $\mathcal{F}_{\text{low}}$, $\mathcal{F}_{\text{mid}}$, $\mathcal{F}_{\text{high}}$, respectively.
Each masked spectrum is then transformed back into the spatial domain using the inverse DFT (see Eq.~\eqref{eq:IDFT}), yielding frequency-specific feature maps for subsequent adaptation.

The resulting spatial-domain features are then processed by dedicated adaptation branches, allowing the model to adjust representations in a frequency-aware manner. Architectural details of the frequency-specific modules are provided in Sec.~\ref{sec:BW}.

\subsection{Band-wise Adaptation} 
\label{sec:BW}
Band-wise Adaptation performs frequency-specific refinement on features isolated by the Frequency Diversion. 
Given that low-, mid-, and high-frequency components encode complementary visual information—from global structure to fine-grained details—uniform processing may hinder adaptation.
To address this, we employ a multi-branch architecture in which each branch is dedicated to adapting features from a specific spectral band.

Let $\{ \mathcal{F}_{\text{low}}, \mathcal{F}_{\text{mid}}, \mathcal{F}_{\text{high}} \}$ denote the frequency-domain representations produced by the Frequency Diversion. 
Each band is first projected back into the spatial domain via the inverse DFT (Eq.~\eqref{eq:IDFT}):
\begin{equation}
    h_{\star} = \text{IDFT}(\mathcal{F}_{\star}), \quad \text{for } \star \in \{ \text{low}, \text{mid}, \text{high} \}.
\end{equation}
The resulting feature maps $h_{\star} \in \mathbb{R}^{W \times H \times C}$ are then processed by independent convolutional branches $\text{Conv}_{\star}$, each specialized for a specific frequency band:
\begin{equation}
    \tilde{h}_{\star} = \text{Conv}_{\star}(h_{\star}), \quad \text{for } \star \in \{ \text{low}, \text{mid}, \text{high} \},
\end{equation}
Here, $\text{Conv}_{\star}$ denotes a $k_{\star} \times k_{\star}$ convolutional layer, where $k_{\star} > 1$ is selected to ensure effective frequency modulation, as convolutional kernels with $k_{\star}=1$ lack nontrivial spectral response (see Appendix for theoretical justification). 
Larger kernels are used for low-frequency branches to better capture global structural patterns, while smaller kernels suffice for high-frequency details.

The low-, mid-, and high-frequency outputs are summed to produce the final adapted representation:
\begin{equation}
    f_\alpha(h) = \tilde{h}_{\text{low}} + \tilde{h}_{\text{mid}} + \tilde{h}_{\text{high}}.
\end{equation}
This structure enables frequency-aware and disentangled adaptation by capturing complementary information at different spectral levels. During meta-testing, only the adapter parameters are updated, while the pre-trained backbone remains frozen.

\section{Experiments}
\label{sec:exper}
\textbf{Dataset.} We evaluate our approach on the \textbf{Meta-Dataset} benchmark~\cite{triantafilloumeta}, which is specifically designed for cross-domain few-shot learning. 
It comprises 13 diverse image classification datasets: ImageNet1k~\cite{russakovsky2015imagenet}, Omniglot~\cite{Lake_Salakhutdinov_Tenenbaum_2015}, FGVC-Aircraft~\cite{Maji2013}, CUB-200-2011~\cite{wah2011caltech}, DTD~\cite{Cimpoi2013}, Quick Draw~\cite{Ha_Eck_2017}, Fungi~\cite{fgvcx2018}, VGG Flower~\cite{Nilsback2008}, Traffic Sign~\cite{Houben2013}, MSCOCO~\cite{Lin2014}, MNIST~\cite{lecun1998gradient}, CIFAR-10~\cite{Krizhevsky_2009}, and CIFAR-100~\cite{Krizhevsky_2009}. 
All images are resized to 84$\times$84 following standard evaluation protocol.

\textbf{Experimental Setup.} We adopt the \textbf{\textit{Multi-Domain Learning (MDL)}} configuration, using the first eight datasets in Meta-Dataset for pretraining (seen domains) and reserving the remaining five exclusively for evaluation (unseen domains). 
Evaluation is performed under two standard settings: \textbf{\textit{Varying-way Varying-shot}}, which follows Meta-Dataset’s official sampling strategy with variable support sizes (up to 100 examples), and \textbf{\textit{Varying-way Five-shot}}, which fixes the number of support examples to five per class, creating a more challenging low-data scenario.

\textbf{Implementation Details.} We adopt ResNet-18 as the backbone and initialize it using a model pretrained with URL~\cite{Li_Liu_Bilen_2021}. During meta-testing, only the task-specific adapter parameters are updated using the Adadelta optimizer\cite{zeiler2012adadelta}. 
Following~\cite{yang2024leveraging}, fine-tuning continues until the support accuracy reaches 99\%, with early stopping triggered shortly after. Learning rates are tuned per domain, and full hyperparameter settings are provided in Appendix.
To align with the spatial granularity of different frequency bands, we employ asymmetric kernel sizes: $k_{\text{low}} = 3$, $k_{\text{mid}} = 3$, and $k_{\text{high}} = 5$. All experiments are conducted on a single NVIDIA RTX 4090 GPU.

% While low-frequency components typically encode global structure, we find that high-frequency signals benefit more from wider receptive fields, likely due to their spatial dispersion. This configuration consistently yields improved generalization across domains.

\textbf{Baselines}
We compare FAD against a range of competitive baselines in both evaluation settings. In the Varying-way Varying-shot setting, we evaluate against URT~\cite{liuuniversal}, FLUTE~\cite{Tian_Wang_Krishnan_Tenenbaum_Isola_2020}, tri-M~\cite{Liu2021triM}, URL~\cite{Li_Liu_Bilen_2021}, T-CNAPS~\cite{bateni2022enhancing}, TSA~\cite{li2022cross}, MOKD~\cite{tian2024mokd}, FES~\cite{wang2024feature}, and ProLAD~\cite{yang2024leveraging}.
For the Varying-way Five-shot setting, we evaluate against S-CNAPS~\cite{bateni2020improved}, SUR~\cite{dvornik2020selecting}, URT, URL, TSA, and TA$^2$-Net~\cite{guo2023task}. 

% Among these, URL, TSA, TA$^2$-Net, FES, and ProLAD adopt the same ResNet-18 backbone as ours for fair comparison.

\section{Results}
\subsection{Comparison with State-of-the-Art Methods in Varying-Way Varying-Shot Setting}
\begin{table}[!t]
    \centering
    \setlength{\tabcolsep}{0.8 mm}
    \caption{Comparison with state-of-the-art methods under \textit{Varying-Way Varying-Shot} setting. Mean accuracy with 95\% confidence intervals are reported. $^{\dag}$ indicates methods use the URL backbone. The best result is highlighted in \textbf{bold}, and the second-best result is \underline{underlined}.}
    \vspace{-0.1in}
    \resizebox{\textwidth}{!}{%
    \begin{tabular}{@{}cccccccccc|c@{}}
    \toprule
    \textbf{Method} &  URT & FLUTE & tri-M & URL$^{\dag}$ &  T-CNAPS & TSA$^{\dag}$ &  MOKD & FES$^{\dag}$& ProLAD$^{\dag}$ & FAD$^{\dag}$  \\
    \midrule
    Mark.   &  ICLR 20' & ECCV 20' & ICCV 21' & ICCV 21' & WACV 22' & CVPR 22' & ICML 24' & ML 24' & AAAI 24' & Ours \\
    \midrule
    ImageNet   & 55.0$\pm$1.1 &  51.8$\pm$1.1 & \underline{58.6$\pm$1.0} & 57.5$\pm$1.1 & 57.9$\pm$1.1  & 57.4$\pm$1.1 & 57.3$\pm$1.1 & 56.2$\pm$1.1 & \textbf{59.3$\pm$1.1} &  58.3$\pm$1.1 \\
    Omniglot   &  93.3$\pm$0.5 & 93.2$\pm$0.5 & 92.0$\pm$0.6 &  94.5$\pm$0.4 & 94.3$\pm$0.4 & 95.0$\pm$0.4 & 94.2$\pm$0.5 & 95.3$\pm$0.4 & \textbf{95.4$\pm$0.4} &  \underline{95.3$\pm$0.4}  \\
    Aircraft   &  84.5$\pm$0.6 & 87.2$\pm$0.5 & 82.8$\pm$0.7 &  88.6$\pm$0.5 & 84.7$\pm$0.5 & 89.3$\pm$0.4 & 88.4$\pm$0.5 & 87.6$\pm$0.8 & \underline{89.7$\pm$0.5} &  \textbf{90.5$\pm$0.4}   \\
    Birds      &  75.8$\pm$0.8 & 79.2$\pm$0.8 & 75.3$\pm$0.8 & 80.5$\pm$0.7 & 78.8$\pm$0.7 & 81.4$\pm$0.7 & 80.4$\pm$0.8 & 79.9$\pm$0.8 & \underline{81.7$\pm$0.8} &  \textbf{81.9$\pm$0.7}  \\
    Textures   &  70.6$\pm$0.7 & 68.8$\pm$0.8 & 71.2$\pm$0.8 &  76.2$\pm$0.7 & 66.2$\pm$0.8 & 76.7$\pm$0.7 & 76.5$\pm$0.7 & 76.2$\pm$0.8 & \textbf{78.6$\pm$0.7}  &  \underline{77.8$\pm$0.7}  \\
    Quick Draw &  82.1$\pm$0.6 & 79.5$\pm$0.7 & 77.3$\pm$0.7 &  81.9$\pm$0.6 & 77.9$\pm$0.6 & 82.0$\pm$0.6 & 82.2$\pm$0.6 & \textbf{83.4$\pm$0.6}  & \underline{82.6$\pm$0.6} &   82.3$\pm$0.6\\
    Fungi      &  63.7$\pm$1.0 & 58.1$\pm$1.1 & 48.5$\pm$1.0 & \underline{68.8$\pm$0.9} & 48.9$\pm$1.2 & 67.4$\pm$1.0 & 68.6$\pm$1.0 & \textbf{69.4$\pm$1.1}   & 66.4$\pm$1.1 &  68.0$\pm$1.0 \\ %
    Flower     &  88.3$\pm$0.6 & 91.6$\pm$0.6 & 90.5$\pm$0.5 &  92.1$\pm$0.5 & 92.3$\pm$0.4 & 92.2$\pm$0.5 & 92.5$\pm$0.5 & 91.9$\pm$0.7  & \textbf{93.4$\pm$0.4} &  \underline{93.2$\pm$0.4}  \\
    \midrule
    Traffic Sign & 50.1$\pm$1.1 & 58.4$\pm$1.1 & 78.0$\pm$0.6 & 63.3$\pm$1.2 & 59.7$\pm$1.1 & 83.5$\pm$0.9 & 64.5$\pm$1.1 & 84.9$\pm$1.0  &  \underline{88.5$\pm$0.9} &  \textbf{89.4$\pm$0.8} \\
    MSCOCO       & 48.9$\pm$1.1 & 50.0$\pm$1.0 & 52.8$\pm$1.1 & 54.0$\pm$1.0 & 42.5$\pm$1.1 & \underline{55.8$\pm$1.1} & 55.5$\pm$1.0 & 54.1$\pm$1.0 &  55.6$\pm$1.1 &  \textbf{56.2$\pm$1.1}\\
    MNIST        & 90.5$\pm$0.4 & 95.6$\pm$0.4 & 96.2$\pm$0.3 & 94.5$\pm$0.5 & 94.7$\pm$0.3 & 96.7$\pm$0.4 & 95.1$\pm$0.4 & \underline{97.1$\pm$0.5} &  97.0$\pm$0.3 &  \textbf{97.7$\pm$0.4}  \\
    CIFAR-10     & 65.1$\pm$0.8 & 78.6$\pm$0.7 & 75.4$\pm$0.8 & 71.9$\pm$0.7 & 73.6$\pm$0.7 & \textbf{80.6$\pm$0.8} & 72.8$\pm$0.8 & 78.1$\pm$0.9 & \underline{80.4$\pm$0.9} &  \underline{80.4$\pm$0.8}\\
    CIFAR-100    & 57.2$\pm$1.0 & 67.1$\pm$1.0 & 62.0$\pm$1.0 & 62.6$\pm$1.0 & 61.8$\pm$1.0 & 69.6$\pm$1.0 & 63.9$\pm$1.0 & 70.4$\pm$1.1 & \underline{71.4$\pm$1.0} &  \textbf{71.8$\pm$1.0} \\
    \midrule
    Aver. Seen    & 76.7 & 76.2 & 74.5 & 80.0 & 75.1 & 80.2 & 80.0 & 79.7 & \underline{80.9} &  \textbf{80.9} \\
    Aver. Unseen  & 62.2 & 69.9 & 72.9 & 69.3 & 66.5 & 77.2 & 70.3 & 76.9  & \underline{78.6} & \textbf{79.1}\\
    Aver. All     & 71.1 & 73.8 & 73.9 & 75.9 & 71.8 & 79.0 & 76.3 & 78.7 & \underline{80.0} & \textbf{80.2} \\
    \midrule
    Aver. Rank    & 8.7 & 7.6 & 7.5 & 5.8 & 7.7 & 3.8 & 5.3 & 4.2 & \underline{2.2} &\textbf{1.8}  \\ 
    \bottomrule
    \label{table:main_exp}
    \end{tabular}%
    } 
\vspace{-0.3in}
\end{table}

Table~\ref{table:main_exp} reports performance under the \textit{Varying-Way Varying-Shot} setting on Meta-Dataset, where models are evaluated across 13 diverse datasets using a ResNet-18 backbone pretrained on 8 source domains. 
FAD achieves the highest mean accuracy and the best average rank (1.8), demonstrating strong and consistent performance across heterogeneous visual distributions.

FAD achieves top accuracy on 6 datasets, including fine-grained domains such as Aircraft and CUB, as well as high-variance domains like MSCOCO, highlighting its capacity to model both detailed and diverse patterns. 
It also performs competitively on texture-centric datasets such as Textures and abstract datasets such as Quick Draw, indicating robustness to varying visual modalities.

Crucially, on unseen domains where generalization is most challenging, FAD achieves a new state-of-the-art average accuracy of 79.1\%, outperforming previous methods by a clear margin. Notable gains on Traffic Sign and MNIST further underscore its robustness to substantial domain shifts.

Incorporating frequency-domain knowledge proves essential for cross-domain few-shot learning, as it enables models to capture domain-specific variations that are often missed by spatial-only approaches. 
By combining frequency diversion with band-wise adaptation, FAD supports targeted feature modulation aligned with spectral semantics, leading to improved generalization under distribution shift.

\subsection{Comparison with State-of-the-Art Methods in Varying-Way Five-Shot Setting}
\begin{table}[!t]
    \centering
    \setlength{\tabcolsep}{2.6 mm}
    \caption{Comparison with state-of-the-art methods under \textit{Varying-Way Five-Shot} setting. Mean accuracy with 95\% confidence intervals are reported. $^{\dag}$ indicates methods use the URL backbone. The best result is highlighted in \textbf{bold}, and the second-best result is \underline{underlined}.}
    \vspace{-0.1in}
    \resizebox{0.9\textwidth}{!}{%
    \begin{tabular}{@{}ccccccc|c@{}}
    \toprule
    \textbf{Method} &  S-CNAPS & SUR & URT & URL$^{\dag}$& TSA$^{\dag}$ & TA$^2$-Net$^{\dag}$ &  FAD$^{\dag}$  \\
    \midrule
    Mark. & CVPR 20'  & ECCV 20' & ICLR 20' & ICCV 21' & CVPR 22' & ICCV 23' &  Ours  \\
    \midrule
    ImageNet   & 47.2$\pm$1.0 & 46.7$\pm$1.0 & 48.6$\pm$1.0 & \textbf{49.4$\pm$1.0} & 48.3$\pm$1.1 & \underline{49.3$\pm$1.0} & \underline{49.3$\pm$1.0}  \\
    Omniglot   & 95.1$\pm$0.3 & 95.8$\pm$0.3 & 96.0$\pm$0.3 & 96.0$\pm$0.3 &  \textbf{96.8$\pm$0.3} & 96.6$\pm$0.2 & \textbf{96.8$\pm$0.2}   \\
    Aircraft   & 74.6$\pm$0.6 & 82.1$\pm$0.6 & 81.2$\pm$0.6 & 84.8$\pm$0.5 &  85.5$\pm$0.5 & \underline{85.9$\pm$0.4} & \textbf{86.0$\pm$0.5}     \\
    Birds         &  69.6$\pm$0.7 & 62.8$\pm$0.9 & 71.2$\pm$0.7 & 76.0$\pm$0.6 & 76.6$\pm$0.6 & \underline{77.3$\pm$0.6} & \textbf{78.4$\pm$0.6}   \\
    Textures   & 57.5$\pm$0.7 & 60.2$\pm$0.7 & 65.2$\pm$0.7 & \textbf{69.1$\pm$0.6} &  68.3$\pm$0.7 & 68.3$\pm$0.6 & \underline{68.9$\pm$0.6}  \\
    Quick Draw & 70.9$\pm$0.6 & \underline{79.0$\pm$0.5} & \textbf{79.2$\pm$0.5} & 78.2$\pm$0.5 &  77.9$\pm$0.6 & 78.5$\pm$0.5 & 78.6$\pm$0.6  \\
    Fungi      & 50.3$\pm$1.0 & 66.5$\pm$0.8 & 66.9$\pm$0.9 & 70.0$\pm$0.8 & \textbf{70.4$\pm$0.8} & 70.3$\pm$0.8 & \textbf{70.4$\pm$0.8}  \\ %
    Flower     & 86.5$\pm$0.4 & 76.9$\pm$0.6 & 82.4$\pm$0.5 & 89.3$\pm$0.4 &  89.5$\pm$0.4 & \textbf{90.0$\pm$0.4} & \underline{89.6$\pm$0.4}  \\
    \midrule
    Traffic Sign & 55.2$\pm$0.8 & 44.9$\pm$0.9 & 45.1$\pm$0.9 & 57.5$\pm$0.8 & 72.3$\pm$0.6 & \underline{76.7$\pm$0.5} & \textbf{81.5$\pm$0.4}  \\
    MSCOCO       & 49.2$\pm$0.8 & 48.1$\pm$0.9 & 52.3$\pm$0.9 & \underline{56.1$\pm$0.8} & 56.0$\pm$0.8 & 56.0$\pm$0.8 & \textbf{57.1$\pm$0.8}  \\
    MNIST        & 88.9$\pm$0.4 & 90.1$\pm$0.4 & 86.5$\pm$0.5 & 89.7$\pm$0.4 & 92.5$\pm$0.4 & \underline{93.3$\pm$0.3} & \textbf{94.1$\pm$0.3}   \\
    CIFAR-10     & 66.1$\pm$0.7 & 50.3$\pm$1.0 & 61.4$\pm$0.7 & 66.0$\pm$0.7 & 72.0$\pm$0.7 & \textbf{73.1$\pm$0.7} & \underline{72.2$\pm$0.6}  \\
    CIFAR-100    & 53.8$\pm$0.9 & 46.4$\pm$0.9 & 52.5$\pm$0.9 & 57.0$\pm$0.9 & \underline{64.1$\pm$0.8} & \underline{64.1$\pm$0.8} & \textbf{64.3$\pm$0.8}   \\
    \midrule
    Aver. Seen    & 69.0 & 71.2 & 73.8 & 76.6 & 76.7 & \underline{77.0} & \textbf{77.3} \\
    Aver. Unseen  & 62.6 & 56.0 & 59.6 & 65.2 & 71.4 & \underline{72.6} & \textbf{73.8} \\
    Aver. All     & 66.5 & 65.4 & 68.3 & 72.2 & 74.6 & \underline{75.3} & \textbf{75.9} \\
    \midrule
    Aver. Rank    & 6.0 & 6.0 & 5.1 & 3.6 & 3.0 & \underline{2.3} & \textbf{1.5}  \\
    \bottomrule
    \label{table:vice_exp}
\end{tabular}%
}
\vspace{-0.2in}
\end{table}

Table~\ref{table:vice_exp} reports results under the \textit{Varying-Way Five-Shot} setting, where limited supervision increases the difficulty of adaptation. FAD achieves the highest overall accuracy (75.9\%) and the best average rank (1.5), outperforming all baselines including TA$^2$-Net~\cite{guo2023task} and TSA~\cite{li2022cross}.

Despite the reduced number of labeled examples, FAD maintains its advantages observed in the Varying-Way Varying-Shot setting, achieving a new state-of-the-art accuracy of 73.8\% on unseen domains.
This result highlights its robustness to distribution shift and the efficacy of frequency-aware adaptation in low-data regimes.
On seen domains, FAD remains competitive, ranking first on fine-grained datasets such as Aircraft and CUB, while also performing strongly on abstract or sparse domains like Quick Draw and Textures.

By aligning adaptation with the spectral structure of input features, FAD enables targeted, frequency-specific modulation that enhances generalization across domains. This capability is especially beneficial in low-data scenarios, where conventional spatial methods often struggle to capture subtle, domain-sensitive variations.

\subsection{Ablation and Analysis}
\subsubsection{Ablation on Frequency Diversion and Band-wise Adaptation}
\begin{table}[t]
    \centering
    \setlength{\tabcolsep}{1 mm}
    \caption{Ablation study of the Frequency Diversion Adapter under the \textit{Varying-Way Five-Shot} setting. BW and FD denote Band-wise Adaptation and Frequency Diversion, respectively.}
    \vspace{-0.1in}
    \resizebox{\textwidth}{!}{
    \begin{tabular}{@{}ccccc|cccccccc|ccccc@{}}
    \toprule
     & Backbone & Adapter & BW & FD & ImNet & Omni. & Air. & Brids & Text. & Q. Draw  &  Fungi & Flower & T. Sign & COCO & MNIST & C10 & C100  \\
    \midrule
     \#1 &$\checkmark$ & & & &  \textbf{49.4}  & 96.0 & 84.8 & 76.0 & 69.1 & 78.2 & 70.0 & 89.3 & 57.5 & 56.1 & 89.7 & 66.0 & 57.0 \\
     \#2 &$\checkmark$ & $\checkmark$  & & & 46.3 & 96.5 & 82.4 & 75.9 & 66.6 & 76.3 & 67.3 & 88.8 & 71.1 & 56.4 & 92.0 & 71.8 & 64.1 \\
     \#3 &$\checkmark$ & $\checkmark$ & $\checkmark$ & & 48.2 & 96.5 & 85.2 & 77.1 & 68.3 & 78.5 & \textbf{70.4} & \textbf{89.6} & 80.5 & 56.5 & 93.4 & 70.5 & 63.5 \\
     \#4 &$\checkmark$ & $\checkmark$ & $\checkmark$ & $\checkmark$ & 49.3 & \textbf{96.8} & \textbf{86.0} & \textbf{78.4} & \textbf{68.6} & \textbf{78.6} & \textbf{70.4} & \textbf{89.6} & \textbf{81.5} & \textbf{57.1} & \textbf{94.1} & \textbf{72.2} & \textbf{64.3} \\
    \bottomrule
    \label{tabel:ablation1}
    \end{tabular}
    }
\vspace{-0.15in}
\end{table}

Table~\ref{tabel:ablation1} presents an ablation study under the \textit{Varying-Way Five-Shot} setting, evaluating the contributions of each component in our framework. Starting from a frozen backbone without adaptation (\#1), we add residual adapters (\#2) to enable lightweight task-specific modulation. 
In \#3, we apply group convolution in the spatial domain without transforming features into the frequency domain, thus performing multi-scale adaptation uniformly across all components. 
The full model (\#4) incorporates Frequency Diversion, which transforms features into the frequency domain, partitions them into predefined spectral bands, and applies branch-specific modulation to each band.

Residual adapters (\#1$\rightarrow$\#2) provide substantial improvements, demonstrating their effectiveness in low-data adaptation. Adding band-wise group convolution (\#2$\rightarrow$\#3) yields further gains by capturing multi-scale representations, though the lack of frequency disentanglement limits its expressiveness. Introducing Frequency Diversion (\#3$\rightarrow$\#4) leads to the highest performance, particularly on frequency-sensitive datasets such as CUB and Traffic Sign. These results highlight the benefit of frequency-aware, band-specific modulation in aligning adaptation with domain-specific spectral characteristics, ultimately improving generalization under distribution shift.

\subsubsection{Ablation on Block-wise Adapter Integration}
\begin{figure}[t]
\centering
\includegraphics[width=1\textwidth]{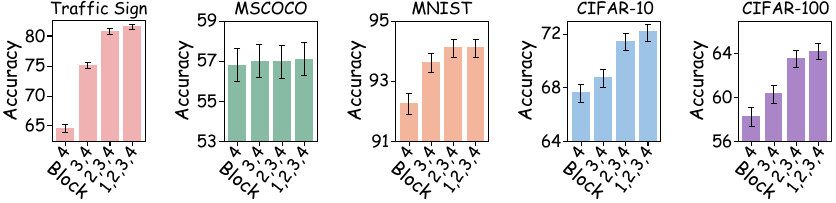}
\vspace{-0.25in}
\caption{Ablation study on block-wise adapter integration under \textit{Varying-Way Five-Shot} setting on unseen domains. Adapters are inserted at different blocks of ResNet-18 to assess the contribution of spectral modulation at various feature depths.} 
\vspace{-0.03in}
\label{fig:ab_block}
\end{figure}

We assess the impact of adapter placement by conducting a block-wise ablation on unseen domains, varying the integration of frequency-aware adapters across the ResNet-18 backbone. As illustrated in Fig.~\ref{fig:ab_block}, inserting adapters into a single block yields limited improvements, whereas applying them across all blocks consistently enhances performance, particularly on Traffic Sign and CIFAR-10.

Access to multi-level features appears essential for robust spectral adaptation. Shallow blocks offer fine-grained texture cues, whereas deeper layers capture more abstract semantics. Incorporating frequency-aware modulation across the full network enables more comprehensive alignment with domain-specific spectral variations, ultimately enhancing generalization under distribution shift.

\subsubsection{Analysis of Kernel Size for Frequency-Specific Modulation}
\begin{table}[t]
    \centering
    \setlength{\tabcolsep}{1.4 mm}
    % \small 
    \caption{Kernel size analysis in Band-wise Adaptation under the \textit{Varying-Way Five-Shot} setting. Each configuration $\{k_{\text{high}}, k_{\text{mid}}, k_{\text{low}}\}$ denotes the use of convolution kernels of size $k \times k$ for the high-, mid-, and low-frequency branches, respectively.}
    \vspace{-0.1in}
    \resizebox{1\textwidth}{!}{%
    \begin{tabular}{@{}ccccccccc|ccccc@{}}
    \toprule
    \diagbox[]{\textbf{Struc.}}{\textbf{Data.}} & ImNet & Omni. & Air. & Brids & Text. & Q. Draw  &  Fungi & Flower & T. Sign & COCO & MNIST & C10 & C100   \\
    \midrule
    \{1, 1, 1\}    &\textbf{49.4} & 95.9 & 85.6 & 77.0 & \textbf{69.4} & 78.2 & 70.4 & 85.6 & 66.8 & 55.8 & 91.1& 66.9 & 59.7   \\
    \{3, 3, 3\}    & 49.3 & 96.3 & 83.7 & 76.0 & 68.9 & 76.7 & 70.4 & 83.7 & \textbf{81.5} & 56.0 & 94.0 & 71.7 & 63.8 \\
    \{3, 3, 5\}    & 49.0& 96.4 & 72.6 & 73.0 & 68.3 & 74.4 & \textbf{70.5} & 72.6  & 77.0  & 51.7 & 92.7 &  67.9 & 57.4 \\
    \midrule
    \{5, 3, 3\}(Ours)& 49.3 & \textbf{96.8} & \textbf{86.0} & \textbf{78.4} & 68.9 & \textbf{78.6} & 70.4 & \textbf{89.5} & \textbf{81.5} & \textbf{57.1} & \textbf{94.1} & \textbf{72.2} & \textbf{64.3}\\
    \bottomrule
    \label{tabel:ablation2}
\end{tabular}%
}
\vspace{-0.25in}
\end{table}

Table~\ref{tabel:ablation2} investigates the role of kernel size in each frequency-specific branch of the adapter. Uniform use of $1 \times 1$ kernels results in consistently degraded performance on unseen domains such as Traffic Sign, MNIST, and CIFAR-10, aligning with Theorem 1 which shows that such kernels lack frequency selectivity. 
Replacing them with $3 \times 3$ kernels leads to consistent improvements, indicating that moderate spatial support is necessary for effective spectral modulation.

The asymmetric configuration $\{5,3,3\}$, which enlarges the receptive field specifically for the high-frequency branch, achieves the best overall performance. This suggests that high-frequency components, which encode fine-grained and spatially dispersed information, benefit from broader spatial aggregation. In contrast, assigning the $5 \times 5$ kernel to the low-frequency branch degrades performance, likely due to oversmoothing of global structures. These findings emphasize the importance of aligning kernel size with spectral granularity, and demonstrate that frequency-aware design is essential for disentangled and effective adaptation.

\subsubsection{Analysis of Band Partition Strategies}
Frequency Diversion (Sec.~\ref{sec:FD}) separates the spectrum into low, mid, and high-frequency bands using radial thresholds $r_1=0.3$ and $r_2=0.5$. 
This design aligns with the inverse power-law distribution observed in natural images, where spectral energy is concentrated near the origin and diminishes toward the periphery. 
The chosen thresholds provide a balanced separation, isolating global structure, mid-level patterns, and fine textures to support semantically aligned modulation across bands.

\begin{wraptable}{r}{0.5\textwidth}
    \centering
    \setlength{\tabcolsep}{1.4mm}
    \vspace{-0.25in}
    \caption{Accuracy under different frequency thresholds $r_1$ and $r_2$ in Frequency Diversion, evaluated on unseen domains in the \textit{Varying-Way Five-Shot} setting.}
    \label{tabel:ana_band}
    % \vspace{-0.1in}
    \resizebox{0.5\textwidth}{!}{%
    \begin{tabular}{@{}ccccc|cccc@{}}
    \toprule
    & \multicolumn{4}{c|}{$r_1 = x$, $r_2 = 0.5$} & \multicolumn{4}{c}{$r_1 = 0.3$, $r_2 = x$} \\
    \cmidrule(lr){2-5} \cmidrule(lr){6-9}
    \multicolumn{1}{r}{$x$} & 0.1 & 0.2 & \textbf{0.3} & 0.4 & 0.4 & \textbf{0.5} & 0.6 & 0.7 \\
    \midrule
    T. Sign & \textbf{83.7} & 82.2 & 81.5 & 80.5 & 81.3 & \textbf{81.5} & 81.3 & 81.3 \\
    COCO    & 56.6 & 56.6 & \textbf{57.1} & 56.5 & 57.0 & \textbf{57.1} & 56.9 & 56.9 \\
    MNIST   & \textbf{94.1} & 94.0 & \textbf{94.1} & 94.0 & 93.9 & \textbf{94.1} & 93.9 & 93.7 \\
    C10     & 71.7 & 71.6 & \textbf{72.2} & 71.7 & 71.8 & \textbf{72.2} & 71.8 & 71.8 \\
    C100    & 63.6 & 63.5 & \textbf{64.3} & 63.6 & 63.9 & \textbf{64.3} & 63.8 & 63.9 \\
    \bottomrule
    \end{tabular}%
    }
    \vspace{-0.15in}
\end{wraptable}
To evaluate the effect of frequency partitioning, we vary the radial thresholds $r_1$ and $r_2$ while preserving the three-band diversion. As shown in Table~\ref{tabel:ana_band}, the configuration $r_1 = 0.3$, $r_2 = 0.5$ achieves the most consistent performance across domains. Varying $r_2$ has limited impact, indicating that the boundary between mid- and high-frequency bands is less sensitive. In contrast, changes to $r_1$ notably affect accuracy, especially on Traffic Sign, where reduced low-frequency coverage improves performance—suggesting greater reliance on high-frequency details. This setting offers a stable partition that preserves global structure while capturing finer textures, facilitating robust, semantically aligned adaptation under domain shift.

\subsection{Visualization of Frequency-Specific Attention Patterns}
\begin{figure}[t]
\centering
\includegraphics[width=\textwidth]{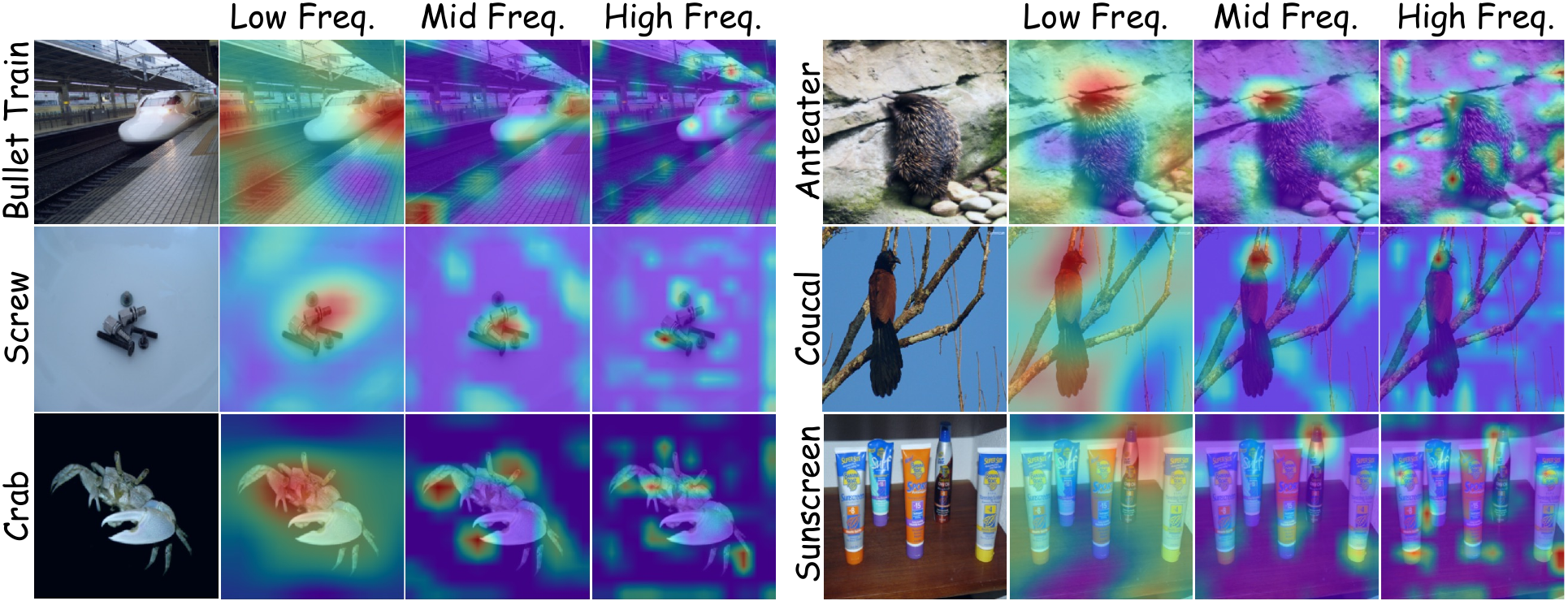}
% \vspace{-0.25in}
\caption{Grad-CAM visualizations for different frequency branches in FAD. For each query image, we visualize the model's attention when only one frequency branch—low, mid, or high—is activated. } 
\vspace{-0.15in}
\label{fig:cam}
\end{figure}
To assess the functional roles of each spectral branch, we visualize Grad-CAM activations from the frequency diversion adapter by selectively activating one frequency branch (low, mid, or high) at a time.
As shown in Fig.~\ref{fig:cam}, attention progressively shifts from coarse to fine spatial granularity: the low-frequency branch captures global structures, the mid-frequency branch highlights semantically salient regions, and the high-frequency branch focuses on localized textures and fine details.

This progressive shift in attention reflects the hierarchical semantic structure embedded in the spectral domain. By enabling band-specific modulation, Frequency Diversion allows the model to capture complementary visual cues at different granularity levels, enhancing its generalization across domains.

\section{Conclusion}
We propose Frequency Adaptation and Diversion (FAD), a frequency-aware framework for cross-domain few-shot learning. 
Central to FAD is the Frequency Diversion Adapter, a lightweight, plug-and-play module compatible with standard backbones. 
Frequency Diversion Adapter integrates Frequency Diversion, which partitions feature representations into spectral bands, with Band-wise Adaptation, which enables semantically aligned, frequency-specific modulation.
FAD achieves state-of-the-art performance on the Meta-Dataset benchmark, demonstrating the effectiveness of incorporating frequency-domain representations for improved generalization.
Ablation studies and visualizations further validate the advantages of spectral partitioning and kernel asymmetry, underscoring the value of frequency-aware adaptation under distribution shift and limited supervision.

\newpage
\bibliography{neurips_2025}
\bibliographystyle{plain}

\end{document}